\documentclass[12pt,oneside,reqno]{amsart}

\usepackage{amssymb,amsthm,amsmath}
\usepackage[foot]{amsaddr}
\usepackage{mathtools} 
\usepackage{graphicx}
\usepackage{amsthm}
\usepackage{bbm}
\usepackage{physics}
\usepackage{enumitem}
\usepackage{float}
\usepackage{tikz}
\usepackage[backref=page, colorlinks=true, allcolors=blue]{hyperref}
\usepackage[abbrev,lite,backrefs]{amsrefs}
\usepackage{color}

\usepackage{multirow}
\usetikzlibrary{shapes}
\usepackage{xparse}
\usepackage{bm} 
\usetikzlibrary{calc,arrows.meta,positioning}
\usepackage{multicol}

\usepackage[colorinlistoftodos,prependcaption,textsize=scriptsize, obeyFinal]{todonotes}

\makeatletter
\providecommand\@dotsep{5}
\renewcommand{\listoftodos}[1][\@todonotes@todolistname]{%
  \@starttoc{tdo}{#1}}
\makeatother

\makeatletter
\def\Ddots{\mathinner{\mkern1mu\raise\p@
\vbox{\kern7\p@\hbox{.}}\mkern2mu
\raise4\p@\hbox{.}\mkern2mu\raise7\p@\hbox{.}\mkern1mu}}
\makeatother

\AtBeginEnvironment{example}{%
  \pushQED{\qed}%
}
\AtEndEnvironment{example}{\popQED\endexample}

\title{Feature Propagation on Knowledge Graphs Using Cellular Sheaves}

\author[J. Cobb]{John Cobb}
\address{Department of Mathematics and Statistics, Auburn University, Auburn, AL}
\email{john.cobb@auburn.edu}

\author[T. Gebhart]{Thomas Gebhart}
\address{Department of Computer Science, University of Minnesota, 200 Union St. SE, Minneapolis, MN 55455 } \email{gebhart@umn.edu}

\thanks{{\em 2020 Mathematics Subject Classification.} Primary: 62R40. Secondary: 55N30, 68T07, 05C50.}

\theoremstyle{plain}
\newtheorem{theorem}{Theorem}[section]
\newcommand{\R}{\mathbb{R}}
\newtheorem{corollary}[theorem]{Corollary}

\newtheorem{remark}{Remark}

\renewcommand{\vec}[1]{{\mathbf{#1}}}
\newcommand{\mat}[1]{{\mathbf{#1}}}

\newcommand{\Slap}{\mat{\Delta}}

\newcommand{\cbdry}{\bm{\delta}}

\newcommand{\Fc}{\mathcal{F}}

\newcommand{\KG}{\mathop{}G}
\newcommand{\Entities}{\mathcal{E}}
\newcommand{\Relations}{\mathcal{R}}
\newcommand{\Triplets}{\mathcal{T}}

\usepackage{xparse}
\usetikzlibrary{calc,arrows.meta,positioning}

\theoremstyle{plain}
\theoremstyle{definition}
\newtheorem{definition}{Definition}
\newtheorem{conjecture}{Conjecture}
\newtheorem{example}[theorem]{Example}

\begin{document}

	
	\begin{abstract}
    Many inference tasks on knowledge graphs, including relation prediction, operate on knowledge graph embeddings -- vector representations of the vertices (entities) and edges (relations) that preserve task-relevant structure encoded within the underlying combinatorial object. Such knowledge graph embeddings can be modeled as an approximate global section of a cellular sheaf, an algebraic structure over the graph. Using the diffusion dynamics encoded by the corresponding sheaf Laplacian, we optimally propagate known embeddings of a subgraph to inductively represent new entities introduced into the knowledge graph at inference time. We implement this algorithm via an efficient iterative scheme and show that on a number of large-scale knowledge graph embedding benchmarks, our method is competitive with--and in some scenarios outperforms--more complex models derived explicitly for inductive knowledge graph reasoning tasks. 
	\end{abstract}
	
	\maketitle 
    \vspace{-0.25in}
	\section{Introduction}
Inferring interactions between nodes in a network is a fundamental and ubiquitous task in relational machine learning. For example, product recommendation via collaborative filtering may be cast as a link prediction problem between customers and products~\cite{koren2009matrix}, and association prediction within a social network corresponds to link prediction between people or communities~\cite{al2006link}. These problems are often defined on knowledge graphs--heterogeneous networks which encode facts about a particular domain by specifying relations (as directed edges) between entities (as nodes).

A common knowledge graph task is to estimate the likelihood of a potential relation existing between two entities, or more generally, a potential logical statement about a group of entities and their interrelationships. The former task is related to simple link prediction (e.g. for \textit{knowledge graph completion}), and the latter to more complex logical reasoning. For example, a simple link prediction task may involve inferring whether a given user has purchased a particular product, or whether a film is set in a particular city. By contrast, a more complex logical reasoning task would involve inferring which user has purchased a basket of products, or the actor starring in a movie set in a given city with a particular writer. 

In the \textit{transductive setting}, wherein all entities and relation types are assumed fixed and identical between training and inference time, there are a wide variety of effective embedding-based methods to perform knowledge graph completion and logical reasoning tasks \cites{bordes2011learning, bordes2013translating, sun2019rotate, zhang2019quaternion}. These transductive knowledge graph embedding methods typically seek to learn a vector-like representation of each entity and each relation type and have seen continued progress and research interest in recent years~\cite{ali2020benchmarking}. 

However in practice, the assumption that entities and relations are known at inference time is often violated; new products are added and removed from inventories, and new films are released over time. Because modern knowledge graphs can contain millions of entities and billions of relational facts, re-learning these transductive representations each time a new entity or relation is introduced--presently the only way for well-known transductive methods like TransE~\cite{bordes2013translating} or Structured Embedding (SE)~\cite{bordes2011learning} to handle the introduction of new entities or relations--is often infeasible. 

This more general environment, wherein the inference knowledge graph may deviate from the training graph, is known as the \textit{inductive setting} and has received considerable attention in recent years~\cites{hamilton2017inductive, teru2020inductive, zhu2021neural, yan2022cycle, geng2022relational}. Many of these approaches employ graph neural networks (GNNs) to learn latent representations of each entity or pair of entities~\cite{zhu2021neural} based on the pattern of relations within some neighborhood of the query. While these approaches are effective for inductive query prediction, the increase in parameter and methodological complexity introduced by these models results in latent entity representations which are highly task-specific and only definable with respect to their integration within a larger knowledge graph, thereby significantly reducing their interpretability and adaptability to other downstream tasks. 

\begin{figure}
    \centering
    \includegraphics[width=0.65\textwidth]{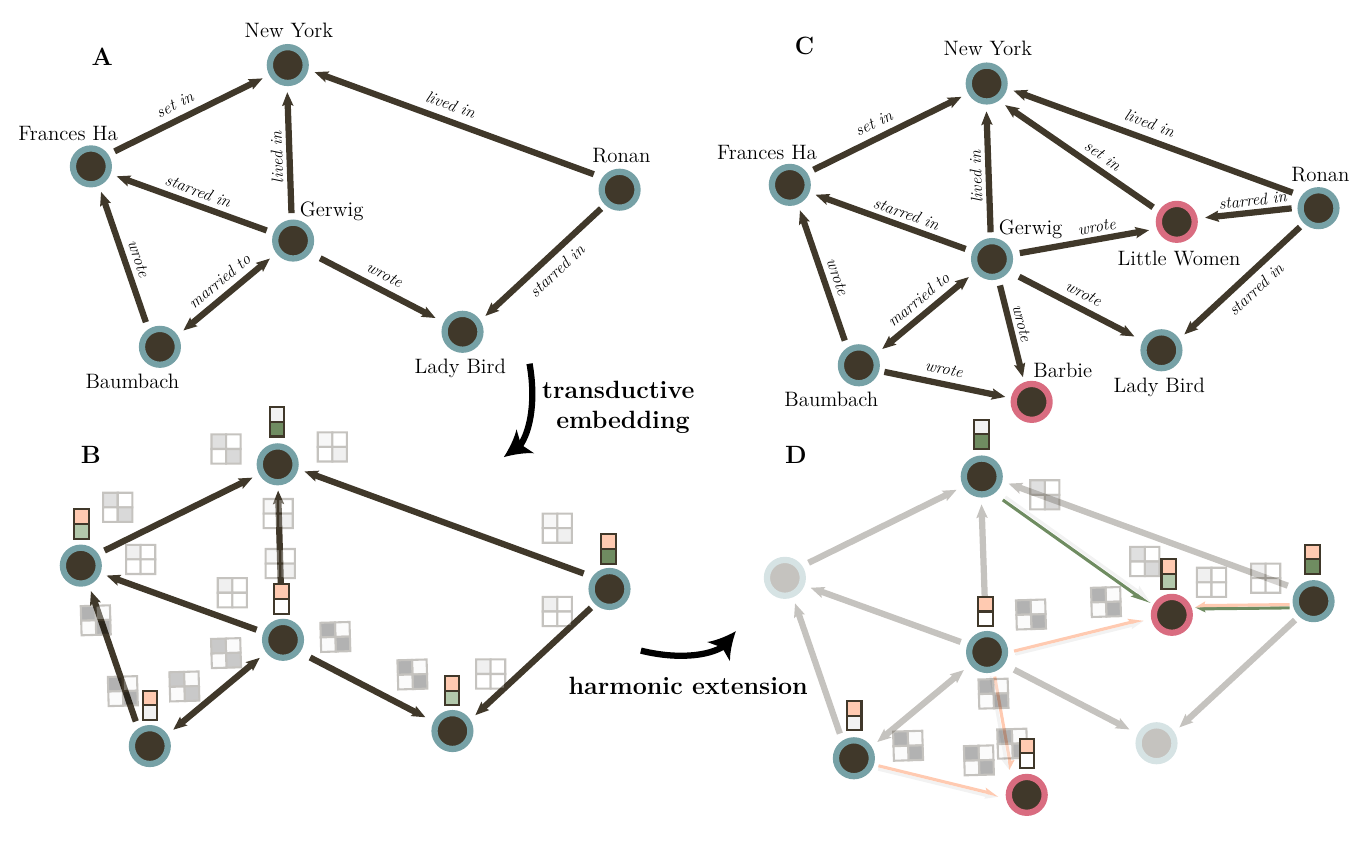}
    \caption{Overview of method. \textbf{A}: An example knowledge graph. $\textbf{B}$: The representations of entities and relations are learned using a knowledge graph embedding method. \textbf{C}: A knowledge graph with new entities (colored red) introduced among existing entities (colored blue). $\textbf{D}$: Representations for new entities are inferred by harmonic extension.}
    \label{fig:inductive_knowledge_graph_extension}
\end{figure}

In this paper, we present a general feature propagation approach for handling missing entity representations for knowledge graph learning tasks. These representations can then be used for downstream tasks on the inference graph, extending a large family of transductive relational learning methods to the inductive setting. In broad strokes, our method leverages the framework of cellular sheaves developed for knowledge graphs in \cite{gebhart2023knowledge} to propagate existing entity representations to unknown entities by minimizing the Dirichlet energy of an associated \textit{sheaf Laplacian} that encodes learned relationships between known entities. This minimization has a closed-form solution given in Theorem \ref{thm:harmonic_extension}, which can be directly computed for small networks and is an optimal harmonic extension given the representations of known entities. We also provide an efficient iterative implementation of this minimization problem in Theorem \ref{thm:training_rate} via gradient descent on this sheaf Dirichlet energy, including training guarantees that ensure the method scales to support harmonic extension over real-world knowledge graphs with more than hundreds of thousands of nodes. In the last section, we examine the efficacy of this process by measuring performance of two popular downstream tasks: knowledge graph completion and logical query reasoning. 

In summary, the method for inductive knowledge graph reasoning presented in this work has the following advantages:

\begin{itemize}[leftmargin = 4 mm]
    \item \textbf{Theoretically Motivated}: Our technique is a natural generalization of harmonic extension and thus comes with optimality guarantees. In particular, our method provably estimates a closed-form solution for the optimal representations of new entities subject to consistency conditions imposed by the relation representations.

    \item {\textbf{Simple and extensible}: Our approach can be directly applied to a large family of popular transductive embedding models, thus providing the means to immediately extend these models to inductive reasoning tasks without retraining.}

    \item \textbf{Robust to high rates of missing data}: Even in the fully-inductive case where \textit{all} entity embeddings are unknown at inference time, performing harmonic extension with randomized entity embeddings is, perhaps surprisingly, still effective. 
\end{itemize}

\subsection*{Acknowledgements} J. Cobb acknowledges the support of the National Science Foundation Grant DMS-2402199.


\section{Background and Notation}\label{sec: Background}
\subsection{Knowledge Graphs and Their Embeddings}\label{sec: Knowledge Graphs}
A \textit{knowledge graph} is a directed multigraph with labeled edges. The nodes are referred to as the set of entities $\Entities$, the labels as relation types $\Relations$, and the labeled directed edges as triplets $\Triplets \subseteq \Entities \times \Relations \times \Entities$ thought of as facts linking the head entity to a tail entity via a relation type. Typically, we think of knowledge graphs as respecting an underlying \textit{knowledge schema} which restricts the possible types of triplets for a given graph according to an entity typing scheme. For instance, we may have a knowledge graph constructed from a social network composed of various types of entities such as people, interest groups, and items. Some relations (e.g. being friends) may make sense between people, but not items.

Broadly speaking, knowledge graph embedding is concerned with finding vectorized representations for each entity and relation such that these representations reflect the epistemic relationships which are encoded by the graph. 
Specifically, these representations are chosen so that the evaluation of some scoring function $f(v,r,w)$ corresponds to the likelihood of a link $(v,r,w)$ being a true fact within the knowledge graph's domain. A wide variety of scoring functions and embedding spaces have been proposed for this task~\cites{bordes2011learning, bordes2013translating, sun2019rotate, zhang2019quaternion}, a select few of which we will introduce and utilize.

Structured Embedding (SE) \cite{bordes2011learning} represents each entity $v$ as a vector $\vec{x}_v \in \mathbb{R}^d$ and each relation $r$ between entities $v$ and $w$ as a pair of $d \times d$ matrices ($\mat{R}_{rv}, \mat{R}_{rw})$. 
It is convenient to think of the transformation $\mat{R}_{rv}$ as embedding $\vec{x}_v$ into a comparison space along an edge as $\mat{R}_{rv}\vec{x}_v$. The scoring function for SE then measures the distance between embeddings of entities along a common edge:
\[ f^{\text{SE}}(v,r,w) = \| \mat{R}_{rv}\vec{x}_v - \mat{R}_{rw} \vec{x}_w \|. \] 

Other embedding models seek to measure the distance between embeddings of entities along common edges \textit{up to some translation}. For instance, TransR \cite{lin2015learning} scores triplets via
\[ f^{\text{TransR}}(v,r,w) = \|\mat{R}_r\vec{x}_v + \vec{r}_r - \mat{R}_r\vec{x}_w \|. \] Here, we think of $\mat{R}_r$ as embedding $\vec{x}_v$ and $\vec{x}_w$ along relation $r$ and $\vec{r}_r$ as a translation associated to $r$. Taking $\mat{R}_r = \mat{I}$ recovers TransE~\cite{bordes2013translating}, which scores a triplet as the discrepancy introduced from translating $\vec{x}_v$ to $\vec{x}_w$ via $\vec{r}_r$: $f^{\text{TransE}}(v,r,w) = \|\vec{x}_v + \vec{r}_r - \vec{x}_w \|$. RotatE~\cite{sun2019rotate} is similar to TransE, but takes entity and relation embeddings to be complex-valued vectors $\vec{x}_v, \vec{r}_r, \vec{x}_w \in \mathbb{C}^d$ with the scoring function $f^{\text{RotatE}}(v,r,w) = \|\vec{x}_v \circ \vec{r}_r - \vec{x}_w \|$ which evaluates the discrepancy introduced via rotation of the representation of $v$ onto the representation of $w$ via $\vec{r}_r$. 

\subsection{Cellular Sheaves}\label{sec:cellular_sheaf_theory}
A sheaf is a mathematical object which tracks the assignment of data along the open sets of a topological space. Using the topology of a graph created from its intrisic partial order of vertices and edges yields a \textit{cellular sheaf}:

\begin{definition}
    A \textit{cellular sheaf} $\Fc$ on an undirected graph $G = (V,E)$ consists of:
    \begin{itemize}
        \item A vector space $\Fc(v)$ for every $v\in V$,
        \item A vector space $\Fc(r)$ for every $r\in E$, 
        \item A linear map $\mat{R}_{rv}: \Fc(v) \to \Fc(r)$ for every incident node-edge pair $v\trianglelefteq r$.
    \end{itemize}
\end{definition}
The vector spaces of the nodes and edge are called \textit{stalks}, while the linear maps are called \textit{restriction maps}. The space formed by all node stalks is called the space of $0$-\textit{cochains} $C^0(G,\Fc) \coloneqq \bigoplus_{v\in V} \Fc(v)$, while the space formed by edge stalks is called the space of $1$-\textit{cochains} $C^1(G,\Fc) \coloneqq \bigoplus_{r\in E} \Fc(r)$.

Given a $0$-cochain $\vec{x} \in C^0(G,\Fc)$, we use $\vec{x}_v$ to mean the vector in $\Fc(v)$ of node $v$. The setup of Section \ref{sec: Knowledge Graphs} can be recast in terms of cellular sheaves \cite{gebhart2023knowledge}: $\vec{x}_v$ plays the role of a potential entity representation of entity $v$, while the choice of cellular sheaf $\Fc$ consists of matrices $\mat{R}_{rv}$ which embeds $\vec{x}_v$ in a comparison space $\Fc(r)$ as $\mat{R}_{rv} \vec{x}_v$. Under this mental framework, the space of \textit{global sections} 
\begin{equation}\label{def: global sections}
    H^0(G,\Fc) \coloneqq \{ \vec{x} \in C^0(G,\Fc) \, | \, \mat{R}_{rw} \vec{x}_w = \mat{R}_{rv} \vec{x}_v\}
\end{equation} 
is the space of entity representations whose embeddings are consistent along each edge with respect to our representation of the given relations. The linear map $\cbdry(\vec{x})_r = \mat{R}_{rw} \vec{x}_w - \mat{R}_{rv} \vec{x}_v$ showing up in (\ref{def: global sections}) is called the \textit{coboundary} map and can be thought to measure the disagreement between entity representations of entity $v$ and $w$ when embedded along relation $r$. The \textit{sheaf Laplacian} aggregates these disagreements together:

\begin{definition}
    The \textit{sheaf Laplacian} of a cellular sheaf $\Fc$ on a graph $G$ is a map $\Slap_\Fc:C^0(G,\Fc) \to C^0(G,\Fc)$ defined as $\Slap_\Fc \coloneqq \cbdry^\top \cbdry$, or node-wise as \[\Slap_\Fc(\vec{x})_w = \sum_{v \stackrel{r}{\to} w} \mat{R}_{rw}^{\top}(\mat{R}_{rw} \vec{x}_w - \
    \mat{R}_{rv} \vec{x}_v).\]
\end{definition}

The sheaf Laplacian is a symmetric positive semi-definite block matrix:
\[ (\Slap_\Fc)_{w,v} = \begin{cases} \sum_{v\trianglelefteq r}\mat{R}_{rv}^\top\mat{R}_{rv} & w=v\\
-\mat{R}_{rw}^\top\mat{R}_{rv} & w\neq v\end{cases}\]
For a trivial sheaf, where all the stalks are $\R$ and the restriction maps are the identity, the sheaf Laplacian simplifies to the graph Laplacian. Cochains such that $\Slap_\Fc(\vec{x}) = 0$ are called \textit{harmonic}. The space of harmonic cochains coincides with the space of global sections:
\begin{equation}\label{eq: ker(L) = H0}
    H^0(G,\Fc) = \ker(\Slap_\Fc).
\end{equation}
When the sheaf $\Fc$ is understood, we use $\Slap$ for short. For more details, we direct the reader to \cite{curry_sheaves_2014} for more information regarding cellular sheaves and to \cite{hansen_toward_2019} for their spectral theory. 

 In this framework, finding a knowledge graph embedding is the same as approximating a cellular sheaf $\Fc$ along with a global section. As in the case of graph Laplacians, repeated applications of the sheaf Laplacian tracks how entity representations diffuse over a larger and larger subset of vertices according to the relation embeddings. More precisely, minimizing the Laplacian quadratic form 
\begin{equation*}
\vec{x}^\top \Slap_{\Fc} \vec{x} = \sum_{v \stackrel{r}{\to} w} \| \mat{R}_{rw} \vec{x}_w - \mat{R}_{rv} \vec{x}_v \|^2  
\end{equation*}
over all possible relation representations and entity embeddings $\vec{x}\in C^0(G,\Fc)$ since, by (\ref{eq: ker(L) = H0}), is equivalent to approximating a sheaf and a global section on the knowledge graph. This intuitively corresponds to finding sections $\vec{x}$ which result from propagating fixed boundary values $\vec{x}_B$ to unknown entities $\vec{x}_U$ in accordance with the restriction maps. 

\section{Harmonic Extension}\label{sec:harmonic_xtension}

In this section we prove our main result (Theorem~\ref{thm:harmonic_extension}), which gives a closed-form solution for the harmonic extension of a knowledge graph embedding from a subgraph and describes an efficient iterative scheme (Theorem~\ref{thm:training_rate}) that converges to this solution.

\subsection{Classical Harmonic Extension} The concept of harmonic extension is a well-studied topic within spectral graph theory. 
Given a weighted graph $G = (V,E,\mat{A})$ with weighted adjacency matrix $\mat{A}$ and degree matrix $\mat{D}$, the graph Laplacian is $\Slap = \mat{D} - \mat{A}$.
For a signal $\vec{x} \in \R^{|V|}$ assigning a scalar to each node, the equation $\Slap \vec{x} = \bm{0}$ is a discrete analogue of Laplace's equation on a manifold without boundary.

To impose Dirichlet boundary conditions, fix a subset $B \subset V$ and write $U = V \setminus B$ for the interior. We seek $\vec{x}$ such that 
\begin{align*}
    \Slap\vec{x}_U &= \vec{0}, \\
    \vec{x}_B &= \vec{y},
\end{align*} 
where $\vec{x}_U$ and $\vec{x}_B$ are the restrictions of $\vec{x}$ to $U$ and $B$. Block-decomposing $\Slap$ with respect to $U\sqcup B$ gives
\begin{equation*}
    \begin{bmatrix}
        \ \Slap[U,U] & \Slap[U,B] \ \\
        \ \Slap[B,U] & \Slap[B,B] \
    \end{bmatrix} 
    \begin{bmatrix}
        \vec{x}_U \\
        \vec{y}
    \end{bmatrix} = 
    \begin{bmatrix}
        \vec{0} \\
        \vec{z}
    \end{bmatrix} 
\end{equation*}
for some free vector $\vec{z}$. Thus $\vec{x}_U$ is determined by 
\[\Slap[U,U]\vec{x}_U + \Slap[U,B] \vec{y} = \vec{0}.\]
Whevenever $\Slap[U,U]$ is invertible (equivalently, every interior vertex has a path to $B$), this linear system has the unique solution 
\begin{equation}\label{eqn:classic_harmonic_extension}
    \vec{x}_U = -\Slap[U,U]^{-1}\Slap[U,B]\vec{y},
\end{equation}
the \textit{harmonic extension} of the boundary value $\bm{y}$ to the interior $U$. 

Equivalently, $\vec{x}$ solves the constrained minimization problem
\begin{equation*}
    \min_{\vec{x}:\vec{x}_B = \vec{y}} \epsilon(\vec{x},G) \coloneqq \vec{x}^T \Slap \vec{x} = \sum_{v\sim w} \mat{A}_{vw}(\vec{x}_v - \vec{x}_w)^2,
\end{equation*}
so that \eqref{eqn:classic_harmonic_extension} is the unique minimizer of the Dirichlet energy subject to the boundary constraints.

\subsection{Harmonic Extension on Knowledge Graphs} We now generalize harmonic extension to the vector valued signals and matrix-valued relations of knowledge graph embeddings. Let $G=(\Entities,\Relations, \Triplets)$ be a knowledge graph with entity embeddings $\{\vec{x}_v\}_{v\in \Entities}$ and relation parameters $\{\mat{R}_{rv}, \mat{R}_{rw}, \vec{r}_r)$ for each edge type $r$. We consider the energy functional

\begin{equation}\label{eq:general_dirichlet_energy}
    E(\vec{x},G)
    =
    \sum_{v \stackrel{r}{\to} w}
    \bigl\|\mat{R}_{rw}\vec{x}_w + \vec{r}_r - \mat{R}_{rv}\vec{x}_v\bigr\|^2,
\end{equation}
where $\vec{x} = \bigoplus_{v \in \mathcal{E}}\vec{x}_v$ is the concatenation of all entity embeddings. The corresponding triplet scoring function is 
\begin{equation}\label{eq:transr}
    f^E(v,r,w)
    =
    \bigl\|\mat{R}_{rw}\vec{x}_w + \vec{r}_r - \mat{R}_{rv}\vec{x}_v\bigr\|^2.
\end{equation}
This family includes several popular models as special cases, e.g. Structured Embedding \cite{bordes2011learning} (if $\vec{r}_r =0$ for all $r$), TransE \cite{bordes2013translating} in $\R^d$, and RotatE \cite{sun2019rotate} in $\mathbb{C}^d$.

Let $\bm{\delta}$ be the sheaf coboundary operator associated to $G$ so that the sheaf Laplacian is $\Slap = \bm{\delta}^\top \bm{\delta}$. We partition the vertex set as $\Entities = U \sqcup B$, where $B$ are the nodes with pretrained embeddings (e.g. training entities), and $U$ are interior nodes to be inferred.

\begin{theorem}\label{thm:harmonic_extension}
Let $G = (\Entities,\Relations,\Triplets)$ be a knowledge graph with relation parameters $\mat{R}_{rv},\mat{R}_{rw},\vec{r}_r$ and entity embeddings $\vec{x}_B$ fixed on a boundary set $B\subset\Entities$.
Assume every interior vertex in $U = \Entities\setminus B$ has a path to $B$, so that $\Slap[U,U]$ is invertible.
Then the unique minimizer of $E(\vec{x},G)$ in~\eqref{eq:general_dirichlet_energy} subject to the boundary constraint $\vec{x}_B$ is given by
\begin{equation}\label{eq:sheaf_harmonic_extension}
    \vec{x}_U^*
    =
    -\Slap[U,U]^{-1}\Slap[U,B]\vec{x}_B
    + \Slap[U,U]^{-1}(\cbdry^\top \vec{r})_U.
\end{equation}
In particular, $\vec{x}_U^*$ minimizes $E(\vec{x},G)$ and the scoring function $f^E$ over all choices of $\vec{x}_U$.
\end{theorem}
\begin{proof}
    We may rewrite \eqref{eq:general_dirichlet_energy} as
    \begin{equation*}
        E(\vec{x},G) = ||\bm{\delta}\vec{x} - \vec{r}||^2 = \vec{x}^\top\Slap \vec{x} - 2\vec{r}^\top \bm{\delta}\vec{x} + \vec{r}^\top \vec{r} 
    \end{equation*}
    where $\vec{r}$ is the concatenation of all the translation components of the relation representations. With $\vec{x}_B$ fixed, this is a strictly convex quadratic function of $\vec{x}_U$ with gradient
    \begin{equation*}
        \nabla_{\vec{x}_U} \tfrac{1}{2}E(\vec{x},G)
    =
    \Slap[U,U]\vec{x}_U + \Slap[U,B]\vec{x}_B - (\cbdry^\top\vec{r})_U.
    \end{equation*}
    Setting the gradient to zero gives the normal equations
\begin{equation*}
    \Slap[U,U]\vec{x}_U^* + \Slap[U,B]\vec{x}_B - (\cbdry^\top\vec{r})_U = 0,
\end{equation*}
and invertibility of $\Slap[U,U]$ yields the closed form \eqref{eq:sheaf_harmonic_extension}.
\end{proof}

Equation (\ref{eq:sheaf_harmonic_extension}) is the \emph{harmonic extension} of $\vec{x}_B$ to $\vec{x}_U$ with respect to the sheaf structure induced by the relation parameters. When $\vec{r}_r = 0$ for all $r$, the solution simplifies to 
\[ \vec{x}_U^* = -\Slap[U,U]^{-1}\Slap[U,B] \vec{x}_B\]
which has the same structural form as the classical harmonic extension \eqref{eqn:classic_harmonic_extension} and minimizes $E(\vec{x},G)$ for the Structured Embedding model. The term $\Slap[U,U]^{-1}(\cbdry^\top \vec{r})_U$ may be understood as a correction induced by the translational components $\vec{r}_r$.

\begin{remark}
    If $\Slap[U,U]^{-1}$ does not exist, we may use its Moore-Penrose pseudoinverse $\Slap[U,U]^\dag$. However, in this case the solution is not unique and the pseudoinverse returns the minimum norm least squares approximation.
\end{remark}

\subsection{Iterative Scheme and Training Rate}

While Theorem~\ref{thm:harmonic_extension} provides a closed-form solution, computing $\Slap[U,U]^{-1}$ costs $O\left((d|U|)^3\right)$ for embedding dimension $d$, which is prohibitive for large $U$. Treating harmonic extension as the steady state of a gradient flow and approximating it by an explicit Euler scheme provides a more practical approach.

Consider the gradient flow 
\begin{equation*}
    \vec{\dot x}(t) = - \nabla_{\vec{x}} \frac{1}{2}E(\vec{x}(t), G)
\end{equation*} with boundary condition $\vec{x}_B(t) \equiv \vec{x}_B$ and initial condition $\vec{x}(0) = \begin{bmatrix} \vec{x}_U(0) \\ \vec{x}_B \end{bmatrix}$. From the computation in the proof of Theorem \ref{thm:harmonic_extension},
\begin{equation*}
    \nabla_\vec{x} \frac{1}{2} E(\vec{x}(t), G) =  \Slap \vec{x}(t) - \cbdry^\top \vec{r},
\end{equation*}
so the flow becomes
\begin{align*}
    \begin{bmatrix}
        \vec{\dot x}_U(t)\\
        \vec{\dot x}_B(t)
    \end{bmatrix}
    &= -\begin{bmatrix}
    \Slap[U,U] & \Slap[U,B]\\ 
        \vec{0} & \vec{0}
    \end{bmatrix}
    \begin{bmatrix}
    \vec{x}_U(t)\\
    \vec{x}_B
    \end{bmatrix}
    + \begin{bmatrix}
        (\cbdry^\top \vec{r})_U\\
        \vec{0}
    \end{bmatrix}
    \\
    &= \begin{bmatrix}
    -\Slap[U,U] \vec{x}_U(t) - \Slap[U,B]\vec{x}_B + (\cbdry^T\vec{r})_U\\
    \vec{0}
    \end{bmatrix} 
\end{align*}

Discretizing in time using Euler's method with step size $h>0$ yields 
\begin{equation}\label{eqn: euler_scheme}
    \vec{x}_U^{(k+1)}
    = (\mat{I} - h\,\Slap[U,U])\vec{x}_U^{(k)}
      - h\,\Slap[U,B]\vec{x}_B
      + h\,(\cbdry^\top \vec{r})_U,
    \qquad
    \vec{x}_B^{(k)} \equiv \vec{x}_B.
\end{equation}
We now quantify its convergence rate to the harmonic extension $\vec{x}_U^*$.

\begin{theorem}\label{thm:training_rate}
Keep the setup and notation of Theorem~\ref{thm:harmonic_extension} and denote
\[
\mu \coloneqq \lambda_{\min}\bigl(\Slap[U,U]\bigr) > 0,
\qquad
L \coloneqq \lambda_{\max}\bigl(\Slap[U,U]\bigr).
\]
Consider the Euler iteration \eqref{eqn: euler_scheme} with step size $h>0$.
\begin{enumerate}
    \item If $0 < h < \frac{2}{L}$, the iteration~\eqref{eqn: euler_scheme} converges linearly to $\vec{x}_U^*$ for every initial condition:
    \begin{equation}\label{eq:rate_general}
        \|\vec{x}_U^{(k)} - \vec{x}_U^*\|_2
        \leq
        \rho(h)^k\|\vec{x}_U^{(0)} - \vec{x}_U^*\|_2,
    \end{equation}
    where
    \[
        \rho(h)
        \coloneqq \max_{\lambda \in \mathrm{spec}(\Slap[U,U])} |1 - h\lambda|
        < 1
    \] and $k$ is the number of iterations.

    \item In particular, if $0 < h \leq \frac{1}{L}$, then
        $\rho(h) = 1 - h\mu$
    and to guarantee
    $\|\vec{x}_U^{(k)} - \vec{x}_U^*\|_2 \leq \varepsilon$ it suffices that
    \[
        k \geq
        \frac{1}{h\mu}
        \log\left(\frac{\|\vec{x}_U^{(0)} - \vec{x}_U^*\|_2}{\varepsilon}\right).
    \]
\end{enumerate}
\end{theorem}
\begin{proof}
The harmonic extension $\vec{x}_U^*$ satisfies
\[
\Slap[U,U]\vec{x}_U^* + \Slap[U,B]\vec{x}_B - (\cbdry^\top\vec{r})_U = 0,
\]
so $\vec{x}_U^*$ is a fixed point of the affine map in~\eqref{eqn: euler_scheme}.  
Let the error at step $k$ be $\vec{e}^{(k)} \coloneqq \vec{x}_U^{(k)} - \vec{x}_U^*$. Subtracting the fixed point relation from \eqref{eqn: euler_scheme} gives
\[
\vec{e}^{(k+1)}
= (\mat{I} - h\,\Slap[U,U])\vec{e}^{(k)},
\]
and hence
\[
\vec{e}^{(k)} = (\mat{I} - h\,\Slap[U,U])^k \vec{e}^{(0)}.
\]

Since $\Slap[U,U]$ is positive semidefinite, we can diagonalize $\Slap[U,U]$ as $\Slap[U,U] = \mat{Q}\Lambda\mat{Q}^\top$ with
$\Lambda = \mathrm{diag}(\lambda_1,\dots,\lambda_m)$, $0<\mu=\lambda_1\leq\cdots\leq\lambda_m=L$, and orthogonal $\mat{Q}$. Then
\[
\vec{e}^{(k)}
= \mat{Q}(\mat{I} - h\Lambda)^k \mat{Q}^\top \vec{e}^{(0)},
\]
so by the triangle inequality
\[
\|\vec{e}^{(k)}\|_2
\leq \|\mat{Q}\|_2\,\|\mat{Q}^\top\|_2\,
     \max_i |1 - h\lambda_i|^k \,\|\vec{e}^{(0)}\|_2
= \rho(h)^k\,\|\vec{e}^{(0)}\|_2.
\]
The condition $0<h<2/L$ implies $|1-h\lambda|<1$ for all $\lambda\in[\mu,L]$, so $\rho(h)<1$, giving~\eqref{eq:rate_general}.

Moreover if $0<h\leq 1/L$, then $1-h\lambda\in[1-hL,1-h\mu]\subseteq[0,1-h\mu]$ for all $\lambda\in[\mu,L]$, and thus $\rho(h) = \max_{\lambda\in[\mu,L]} (1 - h\lambda) = 1 - h\mu$. Solving
$(1-h\mu)^k \leq \varepsilon / \|\vec{e}^{(0)}\|_2$ for $k$ gives the stated iteration bound.
\end{proof}

In practice, we use the \textit{normalized sheaf Laplacian} 
\begin{equation*}
    \widetilde{\Slap} \coloneqq D^{-\frac{1}{2}} \Slap D^{-\frac{1}{2}}
\end{equation*} where $D$ is the block-diagonal of $\Slap$. This corresponds to an isomorphic sheaf with rescaled stalks and has spectrum contained in $[0,2]$, so that $L<2$ and we may safely choose $h=1$. 

\begin{corollary}\label{cor:h_equals_one}
If $L < 2$, then the choice $h=1$ in~\eqref{eqn: euler_scheme} satisfies $0<h<2/L$, so the iteration converges linearly to the harmonic extension $\vec{x}_U^*$ with rate
\[
\|\vec{x}_U^{(k)} - \vec{x}_U^*\|_2
\leq
\rho(1)^k\,\|\vec{x}_U^{(0)} - \vec{x}_U^*\|_2,
\qquad
\rho(1) = \max_{\lambda\in\mathrm{spec}(\Slap[U,U])}|1-\lambda| < 1.
\]
\end{corollary}

\section{Experiments}\label{sec:experiments}
For all knowledge graph embedding tasks, we assume we are given a knowledge graph $\KG = (\Entities, \Relations, \Triplets)$ consisting of all true facts about a domain.
In this work, $\KG$ represents the entirety of a knowledge base like Freebase~\cite{bollacker2008freebase} or Wordnet~\cite{miller1995wordnet}. 
We define two subgraphs from $\KG$. 
The \emph{training graph} $\KG_{\mathrm{train}} = (\Entities_{\mathrm{train}}, \Relations, \Triplets_{\mathrm{train}}) \subset \KG$ encodes knowledge known about the domain at training time. 
The \emph{inference graph} $\KG_{\mathrm{inf}}^{\mathrm{est}} = (\Entities_{\mathrm{inf}}, \Relations, \Triplets_{\mathrm{inf}}^{\mathrm{est}}) \subseteq \KG$ encodes--not necessarily exclusively from $G_{\mathrm{train}}$--knowledge about the domain to be inferred during evaluation. 
Additionally, we assume the existence of the \emph{observable inference graph} $\KG_{\mathrm{inf}}^{\mathrm{obs}} = (\Entities_{\mathrm{inf}}, \Relations, \Triplets_{\mathrm{inf}}^{\mathrm{obs}}) \subseteq \KG_{\mathrm{inf}}^{\mathrm{est}}$ such that $\Triplets_{\mathrm{inf}}^{\mathrm{obs}} \subset \Triplets_{\mathrm{inf}}^{\mathrm{est}}$. 
We may understand $\KG_{\mathrm{inf}}^{\mathrm{obs}}$ as a knowledge graph delineating how new entities $\Entities_{\mathrm{inf}} \setminus \Entities_{\mathrm{train}}$ are related to $\Entities_{\mathrm{train}}$ given the existing structure of $\KG_{\mathrm{train}}$ and $\Relations$.  
Note that $\KG_{\mathrm{train}}$, $\KG_{\mathrm{inf}}^{\mathrm{obs}}$, and $\KG_{\mathrm{inf}}^{\mathrm{est}}$ are all composed by the same relation set $\Relations$, meaning they share the same schema. We run experiments on knowledge graph completion and logical query reasoning across the following settings:

\begin{itemize}[leftmargin = 4 mm]
    \item \textbf{Transductive Setting.} In the traditional transductive setting wherein $\KG_{\mathrm{train}} = \KG_{\mathrm{inf}}^{\mathrm{obs}} \subset \KG_{\mathrm{inf}}^{\mathrm{est}}$, one seeks to learn representations of $\Entities_{\mathrm{train}}$ and relations $\Relations$ which can be leveraged to correctly infer the inference triplets $\Triplets_{\mathrm{inf}}^{\mathrm{est}} \subseteq \Entities_{\mathrm{train}} \times \Relations \times \Entities_{\mathrm{train}}$. 
    
    \item \textbf{Semi-Inductive Setting.} Note that in the transductive setting, we have implicitly assumed that all possible entities to be observed at inference time are contained within $\KG_{\mathrm{train}}$, i.e. $\Entities_{\mathrm{train}} = \Entities$.
    This assumption is restrictive in many applications where new entities may arise at the time of query inference. 
    Relaxing this assumption gives rise to what we will call the \emph{semi-inductive} knowledge graph completion task which allows for the inference graph to contain entities not seen within the training graph such that $\KG_{\mathrm{inf}}^{\mathrm{est}} \supset \KG_{\mathrm{inf}}^{\mathrm{obs}} \supset \KG_{\mathrm{train}}$, $\Entities_{\mathrm{train}} \subset \Entities_{\mathrm{inf}} \subseteq \Entities$, and $\Triplets_{\mathrm{inf}}^{\mathrm{est}} = \Entities_{\mathrm{inf}} \times \Relations \times \Entities_{\mathrm{inf}}$.
    
    \item \textbf{Inductive Setting.} Finally, the assumption that entities of the training graph have any overlap with entities in the inference graph may be excised entirely, giving rise to the \emph{inductive} setting. 
    In this variation, $\Entities_{\mathrm{train}} \cap \Entities_{\mathrm{inf}} = \emptyset$ so that inference over queries $\Triplets_{\mathrm{inf}}^{\mathrm{est}} \subseteq \Entities_{\mathrm{inf}} \times \Relations \times \Entities_{\mathrm{inf}}$ must depend only on abstract structural knowledge of $\KG_{\mathrm{train}}$ and $\Relations$ instead of relying on explicit entity embeddings. 
    Inductive knowledge graph completion tasks have received significant attention as of late, leading to methods like DRUM~\cite{sadeghian2019drum}, NodePiece~\cite{galkin2021nodepiece}, and  GraIL~\cite{teru2020inductive}. 
\end{itemize}

\subsubsection*{Knowledge Graph Completion} This refers to inferring the existence of inference triplets $\Triplets_{\mathrm{inf}}^{\mathrm{est}}$ given knowledge about $\KG_{\mathrm{train}}$. 
Typically, this inference task amounts to scoring the validity of the tuple $(h,r,t) \in \Triplets_{\mathrm{inf}}^{\mathrm{est}}$ given $h$ and $r$ and observing the score assigned to $(h,r,t)$ relative to $(h,r,t')$ for all other $t' \in \Entities_{\mathrm{\inf}}$. 

\subsubsection*{Logical Query Reasoning}

\begin{figure}[b]
    \centering
    \includegraphics[width=0.60\textwidth]{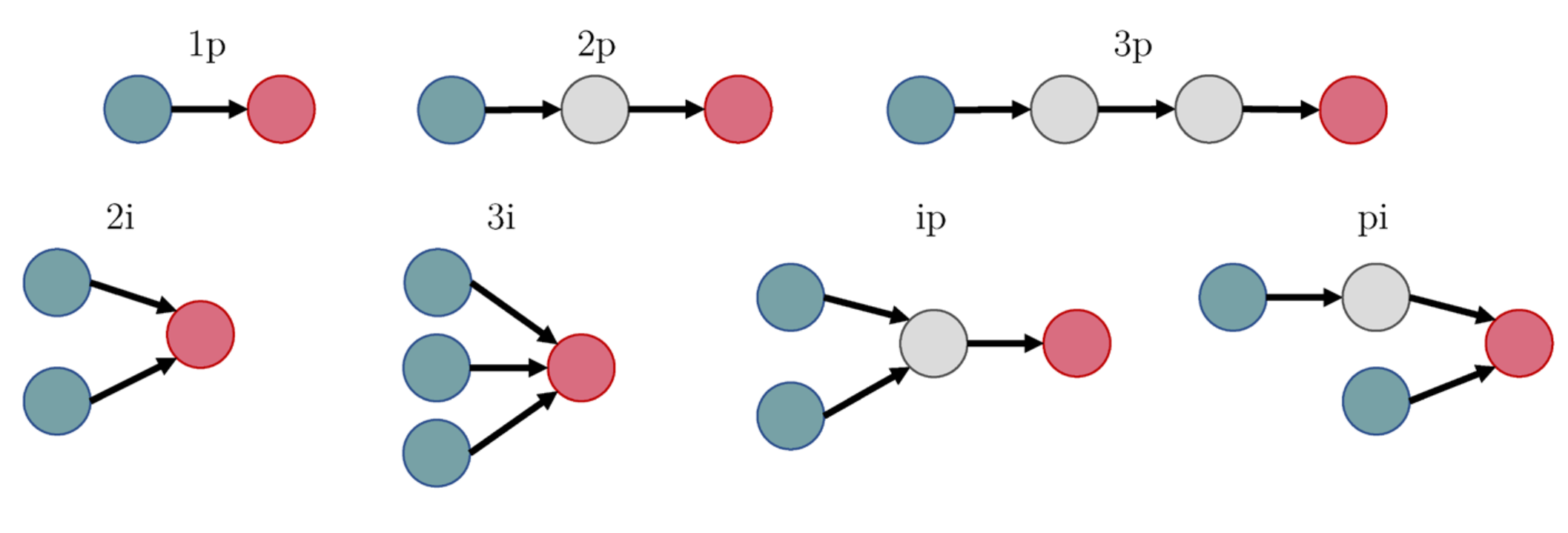}
    \caption{Examples of conjunctive logical query structures considered in this paper. Unknown entities are gray, source entities are colored blue, and target entities are colored red. Evaluating $1p$ queries corresponds to traditional knowledge graph completion.}
    \label{fig:complex_query_structures}
\end{figure}

This task performs reasoning over conjunctive logical query structures like composition ($2p,3p$), conjunction ($2i,3i$), or their combination ($ip,pi)$, as depicted in Figure~\ref{fig:complex_query_structures}. Define a query structure $Q$ as a small subgraph with source vertices $S$ (blue), a target vertex $t$ (red), and interior vertices $U$ (gray). Given embeddings of the sources, the goal of logical query reasoning is to choose a target entity $t$ whose embedding $\vec{x}_t$ is most consistent with $Q$ by evaluating the Dirichlet energy $E(\vec{x},Q)$.

Inference of the entity satisfying a particular triplet $(h,r,?) \in \Triplets_{\mathrm{inf}}^{\mathrm{est}}$ may be viewed as satisfying a simple first-order logical statement $?U:r(h,U)$ where here the relation $r$ is interpreted as a binary logical operation $r: \Entities \times \Entities \to \{\mathrm{true},\mathrm{false}\}$.

By composing the basic relations of a knowledge graph, one can construct more complex, higher-order relationships between entities--for instance, the composition of two \texttt{child of} relations is a new relation \texttt{grandchild} ($?U\exists V: r(h,r(V,U))$.
We refer to these as $*p$ queries. 
Similarly, the intersection of a number of atomic relationships towards a target entity $(h_1, r_1, ?) \wedge (h_2, r_2, ?)$ may be viewed as conjunction $?U\exists V_1, V_2: r_1(V_1, U)\wedge r_2(V_2,U)$. 
We refer to these as $*i$ queries.
We can also combine these projection and conjunction operations to derive $ip$- and $pi$-type queries. 
As shown in Figure~\ref{fig:complex_query_structures}, we can represent these queries as subgraphs $Q$ of the estimation inference graph: $Q \subset \KG_{\mathrm{inf}}^{\mathrm{est}}$. Given a logical query, one seeks to determine whether $Q$ is valid a subgraph of $\KG_{\mathrm{inf}}^{\mathrm{est}}$.
See \cites{ren2020beta, galkin2022inductive, hamilton2018embedding, ren2020query2box} for further details regarding logical reasoning over knowledge graphs.

In our implementation, we effectively eliminate the interior vertices $U$ by a Kron reduction (computed via the Schur complement of the sheaf Laplacian), reducing the query to an effective scoring function depending only on the boundary embeddings $\vec{x}_B = (\vec{x}_S,\vec{x}_t)$. For more details, see \cites{hansen_laplacians_2020, gebhart2023knowledge}. 

\begin{figure*}
    \centering
    \includegraphics[width=.95\textwidth]{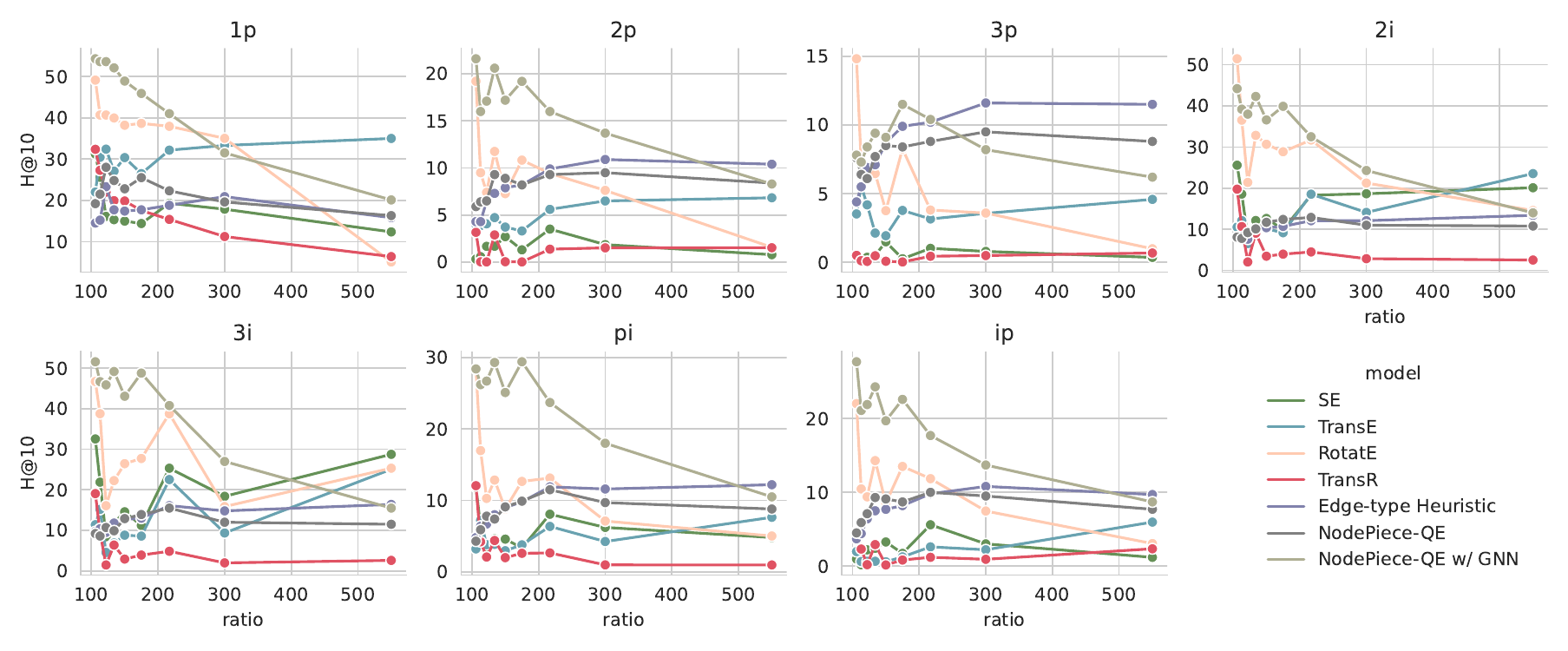}
    \caption{Model performance across different logical query structures with respect to the ratio of inductive to transductive entities. TransE, RotatE, TransR, and SE models are trained transductively on the 1p query completion task then extended to infer representations for new entities. Performance of Edge-type Heuristic and NodePiece models from ~\cite{galkin2022inductive}.}
    \label{fig:transductive_complex_queries}
\end{figure*}

\subsection{Semi-Inductive Logical Query Reasoning}

We use the FB15k-237 logical query reasoning dataset and preparation described in \cite{galkin2022inductive}. 
The data is segmented into nine splits, each corresponding to an increasing ratio of new inference entities relative to the number of entities observed in the knowledge training graph $|\Entities_{\mathrm{inf}} \setminus \Entities_{\mathrm{train}}|/|\Entities_{\mathrm{train}}|$ log-spaced between $[106\%, 550\%]$. 
Each split contains its own training graph $G_{\mathrm{train}}$ along with validation $G_{\mathrm{val}}$ and test $G_{\mathrm{test}}$ inference graphs such that $G_{\mathrm{train}} \subset G_{\mathrm{valid}}$, $G_{\mathrm{train}} \subset G_{\mathrm{test}}$, and $\Entities_{\mathrm{valid}} \cap \Entities_{\mathrm{test}} = \emptyset$.
Each inference graph type in each ratio split additionally contains a number of complex logical queries to be evaluated. 
In this work, we focus on the query types $1p, 2p, 3p, 2i, 3i, pi$, and $ip$ (Figure~\ref{fig:complex_query_structures}). 

For each ratio split, we trained four popular transductive knowledge graph embedding models: TransE, RotatE, TransR, and  SE using the Pykeen~\cite{ali2021pykeen} knowledge graph embedding package.
These transductive models are trained to minimize a consistency loss given only $1p$ queries sampled from $\Triplets_{\mathrm{train}}$.
After optimizing hyperparameters for this transductive task (see Appendix for more information), we re-train the optimal models on $\Triplets_{\mathrm{train}}$ and extend these transductive representations to $\Entities_{\mathrm{valid}}$ by randomly initializing representations for unknown entities in $\Entities_{\mathrm{test}} \setminus \Entities_{\mathrm{train}}$ then performing harmonic extension via the Euler diffusion scheme (Equation~\ref{eqn: euler_scheme}) using entity representations in $\Entities_{\mathrm{train}}$ as the boundary values and relation representations from the transductively-trained models. 
The number of diffusion iterations is determined by the best-performing number of iterations found by running the same extension approach on $\KG_{\mathrm{valid}}$, evaluated on $1p$ queries. 

After extending representations to all of $\Entities_{\mathrm{test}}$, each model was evaluated with respect to each conjunctive logical query structure by ranking all possible tail entities using the Kron reduction of each query subgraph.
The results of this logical query reasoning approach are plotted in Figure~\ref{fig:transductive_complex_queries} using the Hits at 10 (H\@10) retrieval metric. 
As a baseline, the performance of an Edge-type Heuristic model is included which selects all entities in $\Entities_{\mathrm{test}}$ that satisfy the relations in the final hop of the query. 
We also plot the performance of the NodePiece-QE models as described in \cite{galkin2022inductive} which are state-of-the-art models for inductive logical query reasoning. 

Figure~\ref{fig:transductive_complex_queries} indicates that, even though the TransE, RotatE, TransR, and SE models are trained transductively, their functionality can be extended to the semi-inductive setting by inferring new entity representations via harmonic extension, resulting in performance that exceeds the baseline model and occasionally the performance of the state-of-the-art models. 
We find that RotatE and TransE models are most amenable to this extension approach, with RotatE nearing or exceeding state-of-the-art performance on semi-inductive graphs with few new entities versus training entities, and TransE outperforming in higher new-to-training entity ratio environments. 
We also performed hyperparameter tuning of these transductively-trained models based on $pi$ and $ip$ performance in the semi-inductive setting and observe similar or even elevated performance (Appendix). 

\subsection{Inductive Knowledge Graph Completion}

The performance of harmonic extension of transductively-learned embeddings to the semi-inductive setting is influenced by the ratio of the number of new entities to the number of entities present in the training knowledge graph. 
This is expected, as the harmonic extension algorithm detailed in Theorem~\ref{thm:harmonic_extension} assumes the existence of known entity representations to be used as boundary representations anchoring the inference of new entity representations. 
We also investigated the performance of this approach if the overlap between $\Entities_{\mathrm{train}}$ and $\Entities_{\mathrm{test}}$ were removed, and compare performance of this transductive extension method to a number of state-of-the-art models for inductive knowledge graph completion. 

We evaluated model performance on the four version splits of the FB15k-237 and WN18RR datasets detailed in \cite{teru2020inductive}.
We followed the same transductive training and hyperparameter tuning procedure for the TransE, RotatE, TransR, and SE models as was done in the semi-inductive logical query estimation task above. 
Ten percent of the triples in each version's training graph were held out for evaluation during transductive hyperparameter tuning, as the dataset preparation from \cite{teru2020inductive} does not provide transductive validation triplets. 
Relation representations learned during transductive training were retained during inference on $\KG_{\mathrm{test}}$, but entity representations for all of $\Entities_{\mathrm{test}}$ were initialized randomly, corresponding to a truly inductive setting wherein new entities are introduced to the knowledge graph. Treating all entities as interior vertices, we diffuse these randomly-initialized representations according to Equation~\ref{eqn: euler_scheme} then evaluate these diffused representations on $\Triplets_{\mathrm{test}}^{\mathrm{est}}$ for each version split.
We tune the optimal step size $h$ and number of diffusion iterations using $\KG_{\mathrm{valid}}$ of each dataset version split. The performance of a number of state-of-the-art models designed specifically for inductive knowledge graph completion are included for comparison (RuleN~\cite{meilicke2018fine}, Neural-LP~\cite{yang2017differentiable}, DRUM~\cite{sadeghian2019drum}, and GraIL~\cite{teru2020inductive}).

\begin{figure}
    \centering
    \includegraphics[width=0.75\textwidth]{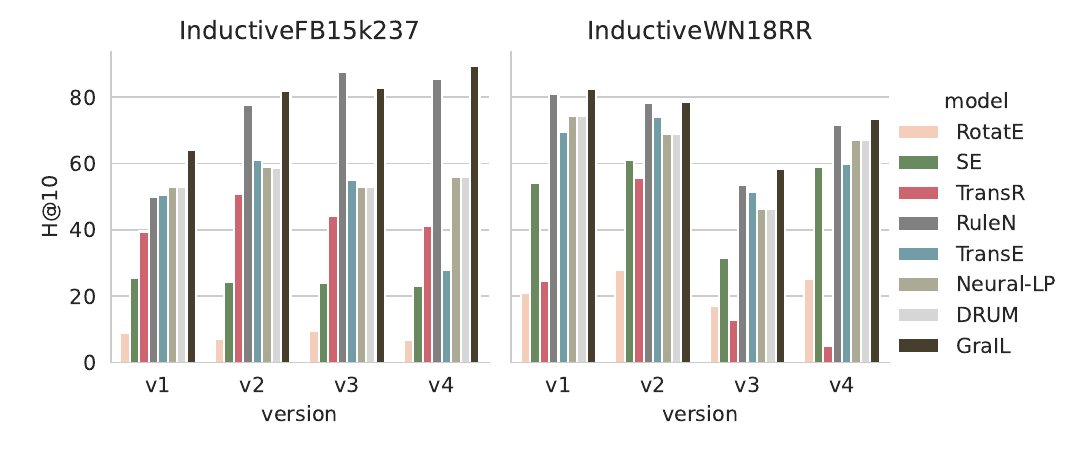}
    \caption{Knowledge graph completion performance in the inductive setting across version splits of each dataset. SE, TransE, RotatE, and TransR models are trained transductively on the knowledge graph completion task then extended to infer entity representations for $\Entities_{\mathrm{test}}$.}
    \label{fig:disjoint_hpo_diff_iterations}
\end{figure}

The results for this inductive knowledge graph completion task are plotted in Figure~\ref{fig:disjoint_hpo_diff_iterations}.
The transductively-extended models perform surprisingly well on this fully-inductive knowledge graph completion task considering their completion inferences are constructed by diffusing the random representations assigned to $\Entities_{\mathrm{test}}$ based solely on their incidence with representations of $\Relations$ learned on $\KG_{\mathrm{train}}$. 
Note that TransE performs in-line with, and occasionally outperforms, Neural-LP and DRUM: two models designed explicitly for this inductive task. 
Although this method for extending transductively-learned representations to the inductive setting presented in this section does not achieve state-of-the-art performance in the fully-inductive setting, this performance implies this approach should serve as a strong and natural baseline against which the performance of inductive-specific model architectures may be compared.

\bibliographystyle{abbrv} 
	\bibliography{final}

\renewcommand{\thesection}{\Alph{section}}

\clearpage
\pagestyle{empty}
\setcounter{section}{0}
\section{Additional Experimental Details}
All experiments were run on a single machine with a single Nvidia A100 GPU with 40GB of RAM. 
Hyperparameter optimization was performed for each model and dataset using the Optuna~\cite{optuna_2019} optimization package. 
We chose loss functions and negative sampling procedures for each model according to the best-performing model configurations found in \cite{ali2020benchmarking}. 
This corresponds to Crossentropy loss without negative sampling for TransE and TransR, and NSSA loss~\cite{sun2019rotate} with negative sampling for RotatE and SE.
All models are optimized using Adam~\cite{kingma2014adam} optimizer without weight decay. 
Remaining hyperparameters (embedding dimension, learning rate, negative sampling ratio, ranking margin and temperature) were optimized based on the models' performance on $1p$ queries in the validation graph.
We used the default optimization search settings and adjusted the number of trials such that each model and dataset completed within 48 hours, capping the maximum number of trials at 200. 
The hyperparameter search configurations and resulting optimal hyperparameters, along with the rest of the implementation can be found at \url{https://github.com/tgebhart/sheaf_kg_transind}. 

\subsection{Choosing Diffusion Iterations}
The number of iterations to run the iterative approximation of harmonic extension represents an important hyperparameter in the inductive inference pipeline. 
Many iterative algorithms employ a stopping criterion, defined with respect to some objective function, to determine when to cease iteration.
However because this stopping criterion is also technically a hyperparameter, we chose instead to optimize for the number of diffusion iterations explicitly to gain a better sense of the relationship between the choice of diffusion iterations and performance. 

\begin{figure}[H]
    \centering
    \includegraphics[width=0.49\textwidth]{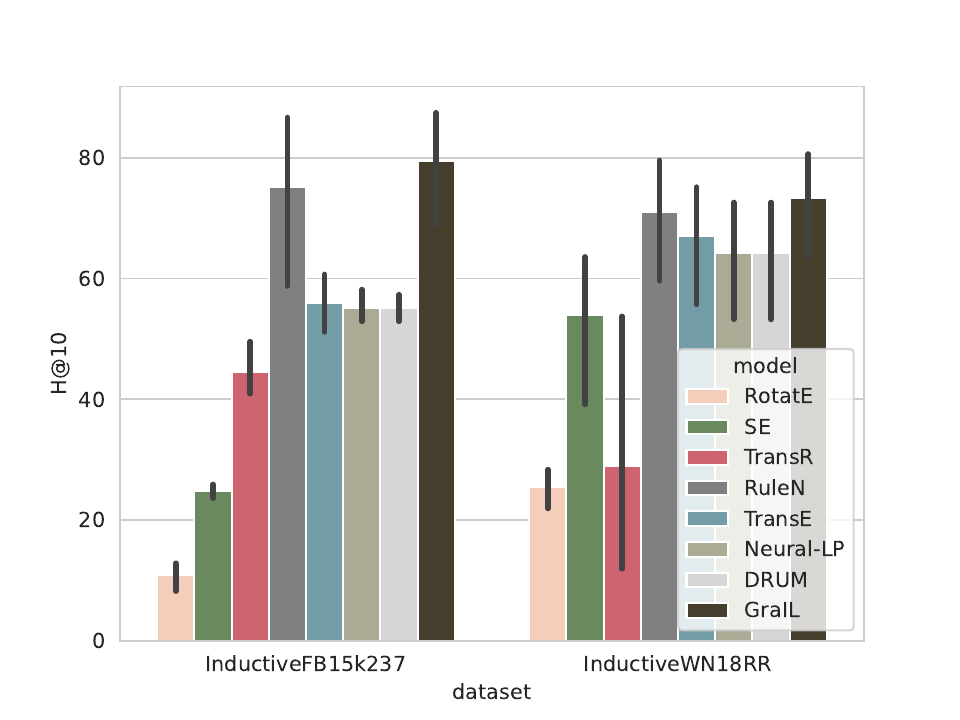}
    \includegraphics[width=0.49\textwidth]{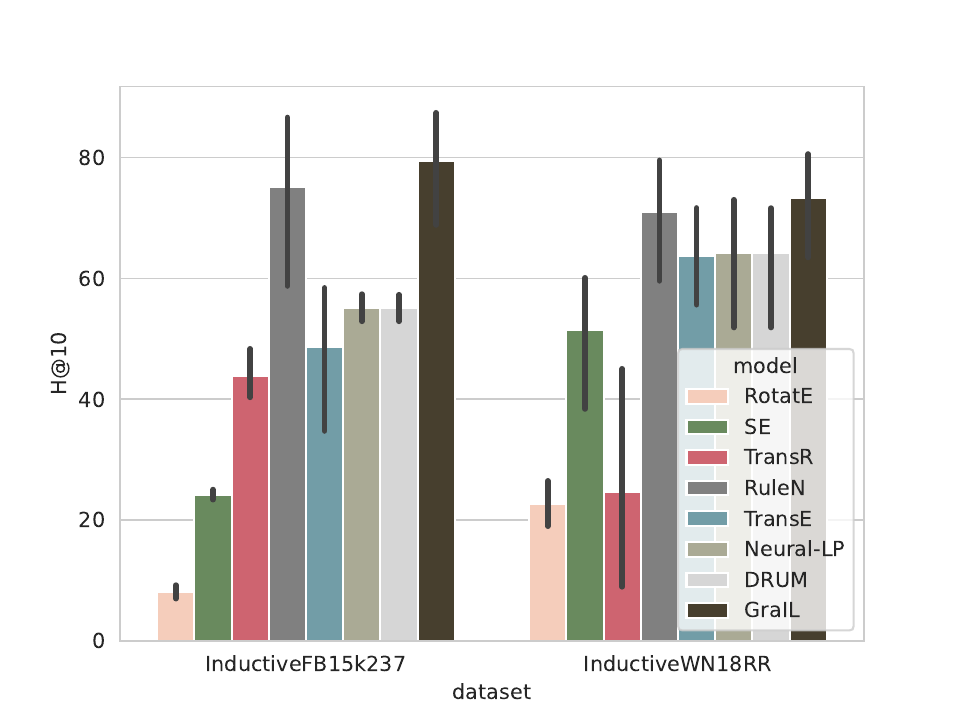}
    \caption{Fully-inductive knowledge graph completion performance with diffusion iterations for SE, TransE, RotatE, and TransR chosen (\textbf{left}) performance is maximized on the test graph and (\textbf{right}) with best performance on the validation graphs Error bars indicate standard error across dataset version splits.}
    \label{fig:overall_max_and_hpo_diff_iterations}
\end{figure}

\begin{figure}[H]
    \centering
    \includegraphics[width=0.75\textwidth]{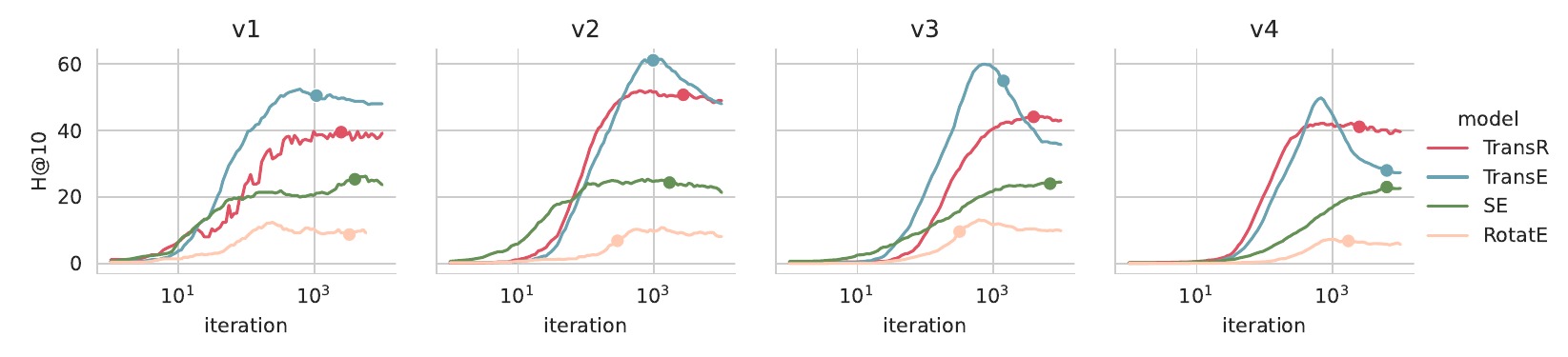}
    \includegraphics[width=0.75\textwidth]{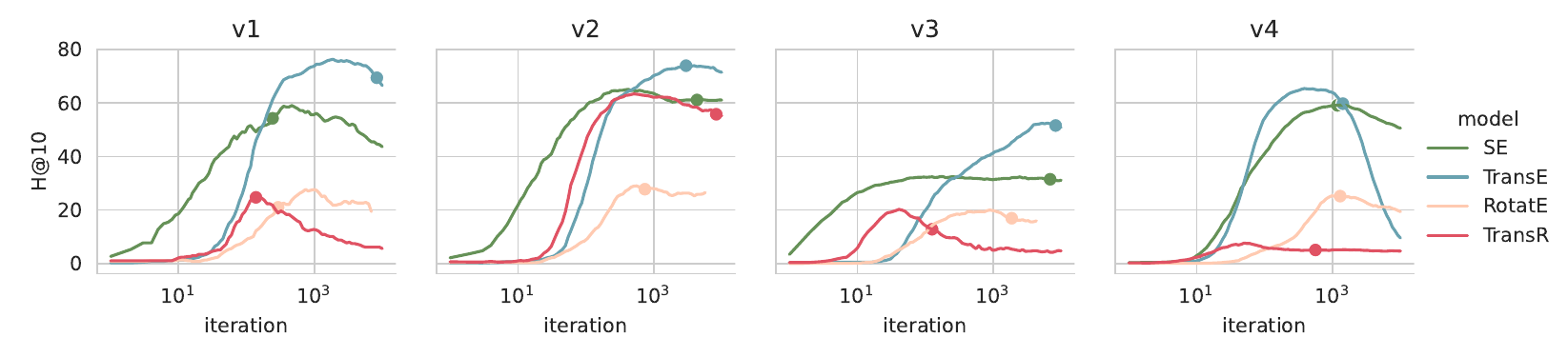}
    \caption{Fully-inductive knowledge graph completion performance on InductiveFB15k237 (top) and InductiveWN18RR (bottom) datasets as a function of diffusion iterations. Plotted point indicates the number of diffusion iterations chosen during hyperparameter optimization on the validation set for each dataset version split.}
    \label{fig:diffusion_iterations}
\end{figure}

The left side of Figure~\ref{fig:overall_max_and_hpo_diff_iterations} plots the performance of these extended models on knowledge graph completion across the InductiveFB15k237 and InductiveWN18RR if we were to use the number of diffusion iterations which results in maximal performance on the test set.
We see the aggregate performance depicted in this plot is only slightly better than that attained when the optimal number of diffusion iterations for the validation set are transferred to the test set, which is plotted on the right of Figure ~\ref{fig:overall_max_and_hpo_diff_iterations}. 
However as can be seen in Figure~\ref{fig:diffusion_iterations}, which plots the performance of each transductive model's performance for knowledge graph completion in the inductive setting on the test set relative to the number of diffusion iterations, the optimal number of diffusion iterations on the validation graph does not always align with the optimal choice for the test graph. 
We leave the determination of more robust methods for choosing the number of diffusion iterations to future work. 

\subsection{Hits@k}
The Hits@k metric is defined as the proportion of true entities with ranking below some threshold $k$:
\begin{equation*}
\mathrm{Hits@K} = \frac{|\{(h, r, t) \in \Triplets_{\mathrm{inf}} \mid \mathrm{rank}_{\Entities_{\mathrm{inf}}}(t) \leq k\}|}
{|\Triplets_{\mathrm{inf}}|}
\end{equation*} where $\mathrm{rank}_{\Entities_{\mathrm{inf}}}(t)$ is the rank of entity $t$'s score in completing the triplet $(h,r,t)$ with respect to all other entities in $\Entities_{\mathrm{inf}}$.

 \begin{figure}[htb]
    \centering
    \includegraphics[height=500pt]{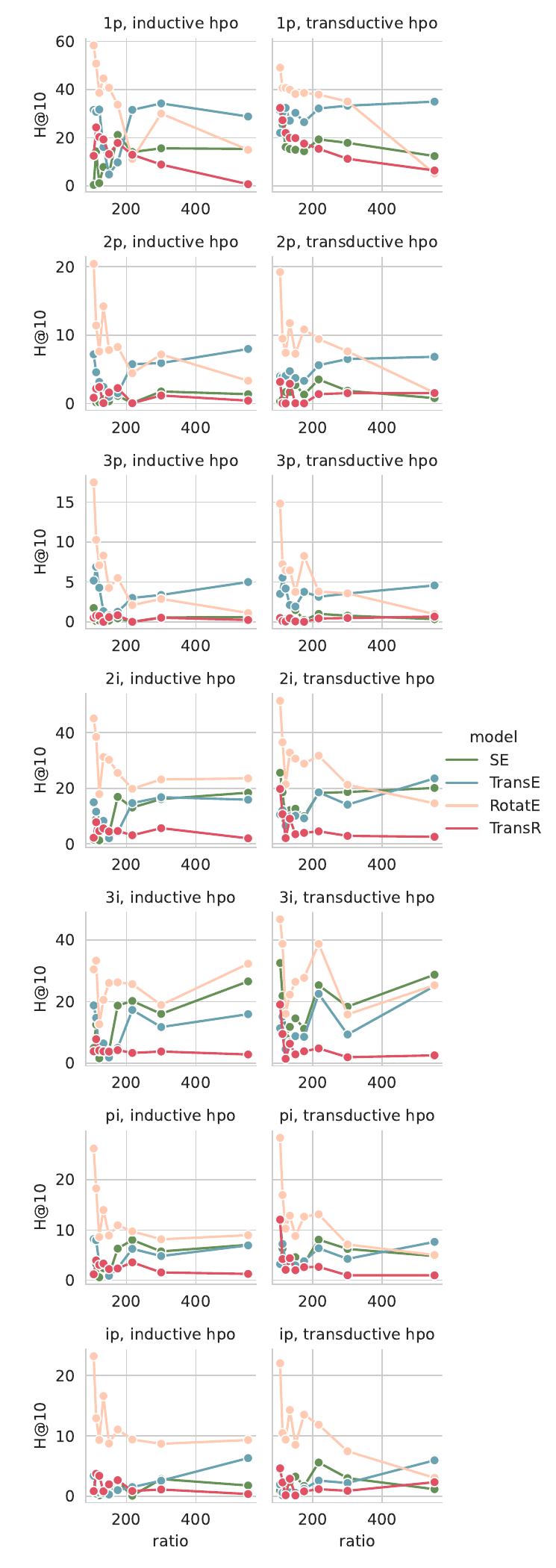}
    \caption{Logical query performance comparison for models with hyperparameters chosen according to inductive performance (inductive) versus hyperparameters chosen according to transductive performance (transductive).}
    \label{fig:transductive_vs_inductive_complex_queries}
\end{figure}
\begin{table}[htbp]
\centering
\footnotesize
\resizebox{0.45\textwidth}{!} {%
\begin{tabular}{@{}cllllllll@{}}
\multicolumn{1}{l}{\textbf{Ratio}} & \textbf{Model}      & \textbf{1p}   & \textbf{2i}   & \textbf{2p}   & \textbf{3i}   & \textbf{3p}   & \textbf{ip}   & \textbf{pi}   \\
\hline \\
\multirow{7}{*}{106}               & Edge-type Heuristic & 14.5          & 8.1           & 4.3           & 9.7           & 4.4           & 3.7           & 4.8           \\
                                   & NodePiece-QE        & 19.2          & 8.1           & 5.9           & 9.2           & 7.6           & 4.5           & 4.3           \\
                                   & NodePiece-QE w/ GNN & \textbf{54.3} & 44.2          & \textbf{21.6} & \textbf{51.6} & 7.8           & \textbf{27.7} & \textbf{28.4} \\
                                   & RotatE              & 49.1          & \textbf{51.5} & 19.2          & 46.7          & \textbf{14.8} & 22.0          & \textbf{28.4} \\
                                   & SE                  & 31.2          & 25.6          & 0.3           & 32.6          & 0.5           & 1.0           & 11.8          \\
                                   & TransE              & 22.0          & 10.5          & 4.0           & 11.4          & 3.5           & 2.0           & 3.2           \\
                                   & TransR              & 32.4          & 19.8          & 3.2           & 19.1          & 0.5           & 4.6           & 12.1          \\
\\ \hline \\                                  
\multirow{7}{*}{113}               & Edge-type Heuristic & 15.2          & 8.2           & 4.3           & 10.4          & 5.5           & 4.4           & 6.4           \\
                                   & NodePiece-QE        & 21.5          & 7.8           & 6.4           & 8.6           & 6.4           & 5.9           & 5.9           \\
                                   & NodePiece-QE w/ GNN & \textbf{53.6} & \textbf{39.2} & \textbf{16.0} & \textbf{46.7} & \textbf{7.3}  & \textbf{21.1} & \textbf{26.2} \\
                                   & RotatE              & 40.7          & 36.5          & 9.5           & 38.8          & 7.2           & 10.5          & 17.0          \\
                                   & SE                  & 25.6          & 18.6          & 0.6           & 21.9          & 0.3           & 0.2           & 6.3           \\
                                   & TransE              & 30.4          & 12.1          & 3.9           & 15.2          & 5.5           & 0.6           & 7.2           \\
                                   & TransR              & 27.3          & 10.7          & 0.0           & 9.5           & 0.1           & 2.3           & 4.2           \\
\\ \hline  \\                                 
\multirow{7}{*}{122}               & Edge-type Heuristic & 23.3          & 7.6           & 6.5           & 9.5           & 6.9           & 6.4           & 6.7           \\
                                   & NodePiece-QE        & 28.0          & 9.2           & 6.5           & 10.7          & 6.1           & 7.1           & 7.8           \\
                                   & NodePiece-QE w/ GNN & \textbf{53.6} & \textbf{38.0} & \textbf{17.1} & \textbf{45.9} & \textbf{8.4}  & \textbf{21.9} & \textbf{26.7} \\
                                   & RotatE              & 40.7          & 21.4          & 7.4           & 16.1          & 6.5           & 9.4           & 10.3          \\
                                   & SE                  & 16.1          & 8.2           & 1.7           & 8.4           & 0.4           & 2.2           & 3.5           \\
                                   & TransE              & 32.4          & 6.5           & 4.1           & 4.6           & 4.2           & 0.6           & 3.8           \\
                                   & TransR              & 22.0          & 2.1           & 0.0           & 1.5           & 0.1           & 0.2           & 2.1           \\
\\ \hline  \\                                 
\multirow{7}{*}{134}               & Edge-type Heuristic & 17.7          & 10.1          & 7.3           & 11.8          & 7.1           & 7.5           & 8.0           \\
                                   & NodePiece-QE        & 24.8          & 10.1          & 9.3           & 9.9           & 7.7           & 9.3           & 7.4           \\
                                   & NodePiece-QE w/ GNN & \textbf{52.1} & \textbf{42.3} & \textbf{20.6} & \textbf{49.2} & \textbf{9.4}  & \textbf{24.3} & \textbf{29.3} \\
                                   & RotatE              & 40.0          & 32.8          & 11.8          & 22.3          & 6.5           & 14.3          & 12.9          \\
                                   & SE                  & 15.3          & 12.1          & 1.7           & 11.8          & 0.5           & 2.4           & 4.6           \\
                                   & TransE              & 27.1          & 9.8           & 4.7           & 6.9           & 2.1           & 0.7           & 3.9           \\
                                   & TransR              & 19.9          & 9.1           & 2.9           & 6.4           & 0.5           & 2.9           & 4.4           \\
\\ \hline \\                                  
\multirow{7}{*}{150}               & Edge-type Heuristic & 17.4          & 10.4          & 7.9           & 12.7          & 8.7           & 7.7           & 9.0           \\
                                   & NodePiece-QE        & 22.8          & 11.7          & 8.9           & 12.9          & 8.5           & 9.1           & 9.1           \\
                                   & NodePiece-QE w/ GNN & \textbf{48.9} & \textbf{36.6} & \textbf{17.2} & \textbf{43.1} & \textbf{9.1}  & \textbf{19.7} & \textbf{25.1} \\
                                   & RotatE              & 38.2          & 30.7          & 7.3           & 26.5          & 3.8           & 8.5           & 8.8           \\
                                   & SE                  & 15.0          & 12.6          & 2.7           & 14.6          & 1.5           & 3.3           & 4.6           \\
                                   & TransE              & 30.4          & 10.1          & 3.7           & 8.8           & 1.9           & 0.6           & 3.0           \\
                                   & TransR              & 19.8          & 3.5           & 0.0           & 2.9           & 0.1           & 0.2           & 2.0           \\
\\ \hline  \\                                 
\multirow{7}{*}{175}               & Edge-type Heuristic & 17.7          & 10.7          & 8.2           & 13.0          & 9.9           & 8.2           & 9.8           \\
                                   & NodePiece-QE        & 25.5          & 12.4          & 8.2           & 13.9          & 8.4           & 8.7           & 9.9           \\
                                   & NodePiece-QE w/ GNN & \textbf{45.9} & \textbf{39.9} & \textbf{19.2} & \textbf{48.8} & \textbf{11.5} & \textbf{22.6} & \textbf{29.4} \\
                                   & RotatE              & 38.6          & 28.9          & 10.8          & 27.7          & 8.3           & 13.5          & 12.7          \\
                                   & SE                  & 14.4          & 10.0          & 1.3           & 11.2          & 0.3           & 1.7           & 3.4           \\
                                   & TransE              & 26.5          & 9.2           & 3.3           & 8.6           & 3.8           & 1.2           & 3.8           \\
                                   & TransR              & 17.5          & 4.0           & 0.0           & 3.9           & 0.0           & 0.8           & 2.6           \\
\\ \hline \\                                  
\multirow{7}{*}{217}               & Edge-type Heuristic & 18.8          & 12.1          & 9.9           & 16.1          & 10.2          & 9.8           & 11.9          \\
                                   & NodePiece-QE        & 22.3          & 12.9          & 9.3           & 15.5          & 8.8           & 10.0          & 11.5          \\
                                   & NodePiece-QE w/ GNN & \textbf{41.0} & \textbf{32.5} & \textbf{16.0} & \textbf{40.8} & \textbf{10.4} & \textbf{17.7} & \textbf{23.7} \\
                                   & RotatE              & 37.9          & 31.7          & 9.4           & 38.8          & 3.8           & 11.9          & 13.1          \\
                                   & SE                  & 19.3          & 18.3          & 3.5           & 25.3          & 1.0           & 5.6           & 8.1           \\
                                   & TransE              & 32.2          & 18.5          & 5.6           & 22.5          & 3.2           & 2.6           & 6.4           \\
                                   & TransR              & 15.4          & 4.5           & 1.4           & 4.8           & 0.4           & 1.2           & 2.7           \\
\\ \hline \\                                  
\multirow{7}{*}{300}               & Edge-type Heuristic & 20.9          & 12.1          & 10.9          & 14.8          & \textbf{11.6} & 10.8          & 11.6          \\
                                   & NodePiece-QE        & 19.6          & 11.0          & 9.5           & 12.0          & 9.5           & 9.5           & 9.7           \\
                                   & NodePiece-QE w/ GNN & 31.5          & \textbf{24.3} & \textbf{13.7} & \textbf{27.0} & 8.2           & \textbf{13.7} & \textbf{18.0} \\
                                   & RotatE              & \textbf{35.0} & 21.2          & 7.6           & 15.9          & 3.6           & 7.5           & 7.1           \\
                                   & SE                  & 17.9          & 18.6          & 1.9           & 18.4          & 0.8           & 3.0           & 6.2           \\
                                   & TransE              & 33.3          & 14.1          & 6.5           & 9.4           & 3.6           & 2.2           & 4.3           \\
                                   & TransR              & 11.2          & 2.9           & 1.5           & 2.0           & 0.5           & 0.9           & 1.0           \\
\\ \hline \\                                   
\multirow{7}{*}{550}               & Edge-type Heuristic & 15.8          & 13.4          & \textbf{10.4} & 16.4          & \textbf{11.5} & \textbf{9.7}  & \textbf{12.2} \\
                                   & NodePiece-QE        & 16.3          & 10.8          & 8.4           & 11.5          & 8.8           & 7.7           & 8.8           \\
                                   & NodePiece-QE w/ GNN & 20.1          & 14.0          & 8.3           & 15.5          & 6.2           & 8.7           & 10.5          \\
                                   & RotatE              & 5.1           & 14.6          & 1.6           & 25.3          & 1.0           & 3.1           & 5.0           \\
                                   & SE                  & 12.4          & 20.1          & 0.8           & \textbf{28.8} & 0.4           & 1.2           & 4.8           \\
                                   & TransE              & \textbf{35.0} & \textbf{23.5} & 6.8           & 25.1          & 4.6           & 6.0           & 7.6           \\
                                   & TransR              & 6.4           & 2.5           & 1.5           & 2.6           & 0.7           & 2.3           & 1.0     \\
\\ \hline
\end{tabular}
}
\caption{Performance of transductive models extended to semi-inductive setting across various percentage ratios of new to existing entities. Models are evaluated on a number of logical query structures. Results for Edge-type Heuristic and NodePiece-QE models from~\cite{galkin2022inductive}. Best-performing models for each ratio and query structure in bold.}
\end{table}

\begin{table}[htbp]
\centering
\footnotesize
\resizebox{0.3\textwidth}{!} {%
\begin{tabular}{@{}ccllll@{}}
\multicolumn{1}{l}{\textbf{dataset}} & \multicolumn{1}{l}{\textbf{version}} & \textbf{model} & \textbf{$\bm{h}$} & \textbf{$\bm{d}$} & \textbf{iterations} \\
\hline \\
\multirow{16}{*}{InductiveFB15k237}  & \multirow{4}{*}{v1}                  & RotatE         & 0.112          & 128                     & 3289                           \\
                                     &                                      & SE             & 0.112          & 256                     & 4076                           \\
                                     &                                      & TransE         & 0.778          & 128                     & 1058                           \\
                                     &                                      & TransR         & 0.334          & 8                       & 2375                           \\
                                     \cline{2-6} \\
                                     & \multirow{4}{*}{v2}                  & RotatE         & 0.889          & 128                     & 306                            \\
                                     &                                      & SE             & 0.778          & 8                       & 1643                           \\
                                     &                                      & TransE         & 0.445          & 128                     & 1014                           \\
                                     &                                      & TransR         & 0.223          & 16                      & 2599                           \\
                                     \cline{2-6} \\
                                     & \multirow{4}{*}{v3}                  & RotatE         & 1              & 256                     & 332                            \\
                                     &                                      & SE             & 1              & 128                     & 6868                           \\
                                     &                                      & TransE         & 0.445          & 128                     & 1416                           \\
                                     &                                      & TransR         & 0.223          & 4                       & 3818                           \\
                                     \cline{2-6} \\
                                     & \multirow{4}{*}{v4}                  & RotatE         & 0.556          & 256                     & 1776                           \\
                                     &                                      & SE             & 0.667          & 256                     & 6352                           \\
                                     &                                      & TransE         & 0.667          & 512                     & 6585                           \\
                                     &                                      & TransR         & 0.112          & 32                      & 2438                           \\
                                   \\  \hline \\
\multirow{16}{*}{InductiveWN18RR}    & \multirow{4}{*}{v1}                  & RotatE         & 0.889          & 128                     & 281                            \\
                                     &                                      & SE             & 1              & 4                       & 240                            \\
                                     &                                      & TransE         & 0.556          & 512                     & 8677                           \\
                                     &                                      & TransR         & 0.445          & 128                     & 134                            \\
                                     \cline{2-6} \\
                                     & \multirow{4}{*}{v2}                  & RotatE         & 1              & 128                     & 723                            \\
                                     &                                      & SE             & 0.445          & 4                       & 4233                           \\
                                     &                                      & TransE         & 0.334          & 64                      & 3122                           \\
                                     &                                      & TransR         & 0.001          & 8                       & 8097                           \\
                                     \cline{2-6} \\
                                     & \multirow{4}{*}{v3}                  & RotatE         & 0.556          & 256                     & 1949                           \\
                                     &                                      & SE             & 0.445          & 16                      & 6781                           \\
                                     &                                      & TransE         & 0.889          & 256                     & 8327                           \\
                                     &                                      & TransR         & 0.112          & 32                      & 123                            \\
                                     \cline{2-6} \\
                                     & \multirow{4}{*}{v4}                  & RotatE         & 1              & 256                     & 1320                           \\
                                     &                                      & SE             & 0.445          & 8                       & 1196                           \\
                                     &                                      & TransE         & 0.556          & 64                      & 1387                           \\
                                     &                                      & TransR         & 0.223          & 128                     & 555  \\                     
                                     \\ \hline 
\end{tabular}
}
\caption{Hyperparameters for transductive models extended to inductive knowledge graph completion task.}
\end{table}
\begin{table}[htbp]
\centering
\footnotesize
\resizebox{0.3\textwidth}{!} {%
\begin{tabular}{@{}clllll@{}}
\multicolumn{1}{l}{\textbf{dataset}} & \textbf{model} & \textbf{v1}   & \textbf{v2}   & \textbf{v3}   & \textbf{v4}   \\
\hline \\
\multirow{8}{*}{InductiveFB15k237}   & DRUM           & 52.9          & 58.7          & 52.9          & 55.9          \\
                                     & GraIL          & \textbf{64.2} & \textbf{81.8} & 82.8          & \textbf{89.3} \\
                                     & Neural-LP      & 52.9          & 58.9          & 52.9          & 55.9          \\
                                     & RotatE         & 8.8           & 7.1           & 9.7           & 6.8           \\
                                     & RuleN          & 49.8          & 77.8          & \textbf{87.7} & 85.6          \\
                                     & SE             & 25.4          & 24.4          & 24.0          & 23.0          \\
                                     & TransE         & 50.5          & 61.1          & 54.9          & 28.0          \\
                                     & TransR         & 39.5          & 50.7          & 44.1          & 41.1          \\
\\ \hline \\
\multirow{8}{*}{InductiveWN18RR}     & DRUM           & 74.4          & 68.9          & 46.2          & 67.1          \\
                                     & GraIL          & \textbf{82.5} & \textbf{78.7} & \textbf{58.4} & \textbf{73.4} \\
                                     & Neural-LP      & 74.4          & 68.9          & 46.2          & 67.1          \\
                                     & RotatE         & 21.0          & 27.8          & 16.9          & 25.2          \\
                                     & RuleN          & 80.9          & 78.2          & 53.4          & 71.6          \\
                                     & SE             & 54.3          & 61.1          & 31.5          & 59.1          \\
                                     & TransE         & 69.4          & 73.9          & 51.6          & 59.7          \\
                                     & TransR         & 24.7          & 55.8          & 12.8          & 5.1       \\
\\ \hline
\end{tabular}
}
\caption{Performance of transductive models extended to inductive setting for splits of InductiveFB15k237 and InductiveWN18RR in the knowledge graph completion task. Performance of DRUM, GraIL, Neural-LP, and RuleN models from~\cite{teru2020inductive}. Best-performing models for each dataset and version in bold.}
\end{table}

\begin{table}[htbp]
\centering
\footnotesize
\resizebox{0.49\textwidth}{!} {%
\begin{tabular}{@{}clllllll@{}}
\multicolumn{1}{l}{\textbf{ratio}} & \textbf{model} & \textbf{embedding\_dim} & \textbf{lr} & \textbf{loss} & \textbf{adversarial\_temperature} & \textbf{margin} & \textbf{num\_negs\_per\_pos} \\
\hline \\
\multirow{4}{*}{106}               & RotatE         & 128                     & 1.3E-03     & nssa          & 1.13                              & 23.09           & 30                           \\
                                   & SE             & 64                      & 5.9E-02     & nssa          & 1.83                              & 3.19            & 19                           \\
                                   & TransE         & 256                     & 1.1E-03     & crossentropy  &                                   &                 &                              \\
                                   & TransR         & 32                      & 1.5E-03     & crossentropy  &                                   &                 &                              \\
\\ \hline \\                                   
\multirow{4}{*}{113}               & RotatE         & 128                     & 1.3E-03     & nssa          & 0.98                              & 25.42           & 57                           \\
                                   & SE             & 64                      & 1.6E-03     & nssa          & 1.33                              & 1.50            & 32                           \\
                                   & TransE         & 32                      & 1.2E-03     & crossentropy  &                                   &                 &                              \\
                                   & TransR         & 64                      & 1.4E-03     & crossentropy  &                                   &                 &                              \\
\\ \hline \\                                   
\multirow{4}{*}{122}               & RotatE         & 256                     & 1.8E-03     & nssa          & 0.19                              & 28.07           & 69                           \\
                                   & SE             & 32                      & 6.1E-02     & nssa          & 1.03                              & 1.01            & 17                           \\
                                   & TransE         & 64                      & 1.1E-03     & crossentropy  &                                   &                 &                              \\
                                   & TransR         & 64                      & 4.5E-03     & crossentropy  &                                   &                 &                              \\
\\ \hline \\                                   
\multirow{4}{*}{134}               & RotatE         & 128                     & 4.1E-03     & nssa          & 0.33                              & 22.63           & 63                           \\
                                   & SE             & 32                      & 7.4E-02     & nssa          & 1.34                              & 2.49            & 50                           \\
                                   & TransE         & 32                      & 1.0E-03     & crossentropy  &                                   &                 &                              \\
                                   & TransR         & 64                      & 1.2E-03     & crossentropy  &                                   &                 &                              \\
\\ \hline \\                                   
\multirow{4}{*}{150}               & RotatE         & 128                     & 4.2E-03     & nssa          & 1.55                              & 21.25           & 28                           \\
                                   & SE             & 32                      & 5.0E-02     & nssa          & 1.48                              & 1.86            & 9                            \\
                                   & TransE         & 128                     & 1.1E-03     & crossentropy  &                                   &                 &                              \\
                                   & TransR         & 64                      & 5.5E-03     & crossentropy  &                                   &                 &                              \\
\\ \hline \\                                   
\multirow{4}{*}{175}               & RotatE         & 128                     & 1.1E-02     & nssa          & 0.04                              & 24.13           & 84                           \\
                                   & SE             & 64                      & 9.6E-02     & nssa          & 0.40                              & 1.30            & 17                           \\
                                   & TransE         & 32                      & 1.0E-03     & crossentropy  &                                   &                 &                              \\
                                   & TransR         & 64                      & 4.6E-03     & crossentropy  &                                   &                 &                              \\
\\ \hline \\                                   
\multirow{4}{*}{217}               & RotatE         & 64                      & 2.6E-02     & nssa          & 0.03                              & 24.17           & 88                           \\
                                   & SE             & 128                     & 2.1E-02     & nssa          & 1.57                              & 1.31            & 7                            \\
                                   & TransE         & 256                     & 1.0E-03     & crossentropy  &                                   &                 &                              \\
                                   & TransR         & 64                      & 7.4E-03     & crossentropy  &                                   &                 &                              \\
\\ \hline \\                                   
\multirow{4}{*}{300}               & RotatE         & 32                      & 5.6E-02     & nssa          & 0.20                              & 10.80           & 49                           \\
                                   & SE             & 64                      & 1.9E-03     & nssa          & 1.94                              & 1.73            & 50                           \\
                                   & TransE         & 128                     & 3.4E-03     & crossentropy  &                                   &                 &                              \\
                                   & TransR         & 32                      & 1.1E-02     & crossentropy  &                                   &                 &                              \\
\\ \hline \\                                   
\multirow{4}{*}{550}               & RotatE         & 256                     & 5.1E-02     & nssa          & 0.41                              & 7.92            & 58                           \\
                                   & SE             & 64                      & 4.1E-02     & nssa          & 1.12                              & 2.00            & 36                           \\
                                   & TransE         & 64                      & 2.1E-02     & crossentropy  &                                   &                 &                              \\
                                   & TransR         & 16                      & 7.5E-02     & crossentropy  &                                   &                 &           \\
\\ \hline                                   
\end{tabular}
}
\caption{Hyperparameters for transductive models extended to semi-inductive logical reasoning task.}
\end{table}
\end{document}